\documentclass[twoside,11pt]{article}

\usepackage[accepted]{melba}

\usepackage{amsmath,amsfonts}

\usepackage{subcaption}
\usepackage{multirow}
\usepackage{amssymb}
\usepackage{bm}
\usepackage{subcaption}
\usepackage{hyperref}
\usepackage[hyphenbreaks]{breakurl}

\usepackage{xcolor}

\melbaid{2023:011}  
\doi{https://doi.org/10.59275/j.melba.2023-3462}
\melbaauthors{Dalca and Sabuncu}  
\volume{2}
\firstpageno{312}  
\melbayear{2023}  
\datesubmitted{08/2022}  
\datepublished{06/2023}  

\melbaspecialissue{Medical Imaging with Deep Learning (MIDL) 2020}
\melbaspecialissueeditors{Marleen de Bruijne, Tal Arbel, Ismail Ben Ayed, Hervé Lombaert}

\ShortHeadings{Multi-task Adversarial CNN Training}{Graziani et al.}
\firstpageno{312}

\title{Learning Interpretable Microscopic Features of Tumor by Multi-task Adversarial CNNs {to Improve} Generalization}

\author{\name Mara Graziani* \email mara.graziani@hevs.ch\\  
	\addr University of Applied Sciences Western Switzerland (HES-SO Valais), 3960, Sierre, Switzerland\\
	\addr University of Geneva (UNIGE), Department of Computer Science (CUI), 1227, Carouge, Switzerland
	\AND
	\name Sebastian Ot\'{a}lora \email juan.otaloramontenegro@hevs.ch \\
		\addr University of Applied Sciences Western Switzerland (HES-SO Valais), 3960, Sierre, Switzerland\\
	\addr University of Geneva (UNIGE), Department of Computer Science (CUI), 1227, Carouge, Switzerland
	\AND 
	\name St\'{e}phane Marchand-Maillet\\
	\addr University of Geneva (UNIGE), Department of Computer Science (CUI), 1227, Carouge, Switzerland
	\AND
    \name Henning M\"uller \email henning.mueller@hevs.ch \\
    \addr University of Applied Sciences Western Switzerland (HES-SO Valais), 3960. Sierre, Switzerland\\
	\addr University of Geneva (UNIGE), Department of Radiology and Medical Informatics, 1211, Geneva, Switzerland
    \AND 
    \name Vincent Andrearczyk \email vincent.andrearczyk@hevs.ch\\
    \addr University of Applied Sciences Western Switzerland (HES-SO Valais), 3960, Sierre, Switzerland\\
}

\begin{document}

\maketitle

\begin{abstract}
Adopting Convolutional Neural Networks (CNNs) in the daily routine of pathological diagnosis requires not only near-perfect precision but also sufficient generalization to data shifts and transparency. 
Existing CNN models act as black boxes, not ensuring the physicians that important diagnostic features are used by the model.
Building on top of successfully existing techniques such as multi-task learning, domain adversarial training and concept-based interpretability, we address the challenge of introducing diagnostic factors in the training objectives.
Our architecture, by learning end-to-end an uncertainty-based weighting combination of multi-task and adversarial losses, is encouraged to focus on pathology features such as density and pleomorphism of nuclei, e.g. variations in size and appearance, while discarding misleading features such as staining variability and acquisition domain.
Our results on breast lymph node tissue show significantly improved generalization in the detection of tumorous tissue, with the best average AUC $0.89 \pm 0.01$ against the baseline AUC $0.86 \pm 0.005$. 
By applying the interpretability technique of linearly probing intermediate representations, we also demonstrate that interpretable pathology features such as nuclei density are learned by the proposed CNN architecture, confirming the increased transparency of this model.
This result is a starting point toward building transparent multi-task architectures that are robust to data heterogeneity.
Our code is available at~\url{ https://github.com/maragraziani/multitask_adversarial}.
\end{abstract}

\begin{keywords}
	Interpretable Deep Learning, Histopathology, Multi-task learning
\end{keywords}

\section{Introduction}
The analysis of microscopic tissue images by Convolutional Neural Networks (CNNs) is an important part of computer-aided systems for cancer detection, staging and grading~\citep{litjens2017survey,janowczyk2016deep,campanella2019clinical,ilse2020deep}. 
The automated suggestion of Regions of Interest (RoIs) is one task that may help pathologists in increasing their performance and inter-rater agreement in the diagnosis \citep{wang2016deep}.
Hard annotations of tumor regions, however, are costly and rarely pixel-level precise. 
Moreover, the existing image datasets are highly heterogeneous, being subject to staining, fixation, slicing variability, multiple scanner resolutions, artifacts, and, at times, permanent ink annotations~\citep{lafarge2017domain}.
While physicians can naturally adapt to the variability of the images in the datasets, deep features are sensitive to confounding factors and their reliability for clinical use is thus still questionable~\citep{janowczyk2016deep,campanella2019clinical}. 
The multiple sources of variability and the limited availability of training data lead to deep features that often present unwanted biases and that do not always mirror clinically relevant diagnostic features. 
For example, using ImageNet for model pretraining introduces an important bias towards texture features~\citep{geirhos2018imagenet} and this impacts the classification of tumorous tissue by transfer learning \citep{graziani2018regression}. 

As largely argued in the literature~\citep{caruana2015intelligible,rudin2019stop,tonekaboni2019clinicians}, transparent and interpretable modelling should be prioritized to ensure model safety and reliability, but transparent models are still struggling to appear in the context of digital pathology. 
This friction is due to the complexity of microscopy images as opposed to working with tabular data with specific clinical descriptions for each variable. 
We thus propose a novel convolutional architecture that increases the transparency and control of the learning process. 
The main technical innovation here is the combination of two successful techniques, namely multi-task learning \citep{caruana1997multitask} and adversarial training \citep{ganin2016domain}, with the purpose of guiding model training to focus on relevant features, called \textit{desired targets}, and to discard \textit{undesired targets} such as confounding factors. 
By experimenting with these techniques, we bring new insights on balancing multiple tasks for digital pathology, that is still an under-explored field~\citep{gamper2020multi}. 
The joint optimization of main, auxiliary and adversarial task losses is a novel exploration in the histopathology field.  


We propose an application of this architecture to the histopathology task of breast cancer classification. 
In this particular context, CNNs should extract from the vast amount of information contained in Whole Slide Images (WSIs) fine-grained features of the tissue structure. 
Our objective function is thus built in such a way to obtain features representative of highly informative clinical factors (e.g. nuclei neoplasticity) and invariant to confounding concepts (e.g. staining variability). 
Our results show that the learned features contain information about the desired targets of nuclei density, area and texture, which are chosen to match the diagnostic procedure of physicians. 
{At the same time, we achieve improved robustness on datasets with distributional shift  given by multiple acquisition centers. }
{As the learning functions for the multiple targets require different loss functions, we bring insights on how intrinsically different tasks can be balanced together in a cumulative loss function. 
The main challenge is, in fact, the combination of losses that have different error metrics such as mean squared error and cross entropy. 
For this reason, we investigate the impact of a dynamic task re-weighting technique based on the uncertainty estimation of each task during training~\citep{kendall2018multi}, which is designed on purpose to facilitate the joint optimization of classification and regression objectives.}
From our analysis, it emerges that this uncertainty-based approach best handles the convergence and stability of the joint optimization.  
Our results also show a significant increase in the performance and generalization to unseen data.
The results are reported in Section~\ref{sec:results} and discussed in Section~\ref{sec:discussion}. 
Further details on experimental setup and methods are presented in Section~\ref{sec:methods}

\section{Related Works}
\label{sec:relatedwork}
\subsection{Multi-task Learning}
\begin{figure}[t!]
    \centering
    \begin{subfigure}{0.32\linewidth}
        \includegraphics[width=\linewidth]{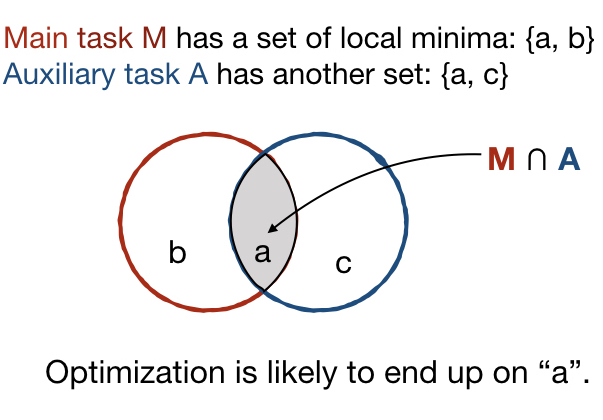}
        \caption{}
        \label{fig:staining}
    \end{subfigure}
    \hfill
    \begin{subfigure}{0.32\linewidth}
        \includegraphics[width=\linewidth]{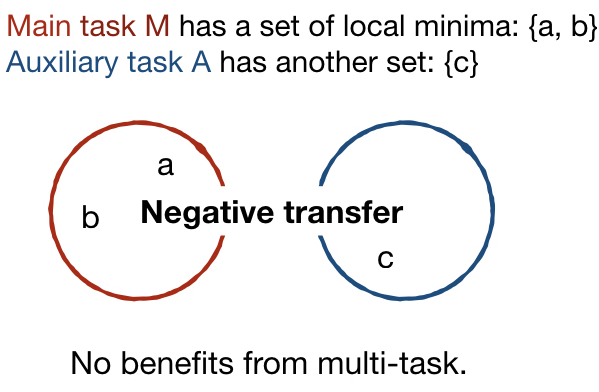}
        \caption{}
        \label{fig:part2}
    \end{subfigure}
    \begin{subfigure}{0.32\linewidth}
        \includegraphics[width=\linewidth]{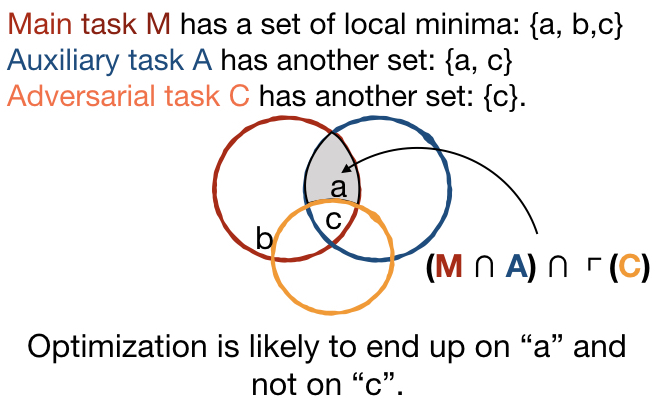}
        \caption{}
        \label{fig:part3}
    \end{subfigure}
    \caption{Intuitive illustration about multi-task learning in (a): given two related tasks M and A, the optimization process is driven to choose solutions that satisfy both tasks. 
    In (b) no connection exists between the tasks, hence the multi-task approach may result in negative transfer, providing only sub-optimal models for all the tasks. 
    In (c), an adversarial task is added and the optimization is pushed to representations that satisfy both main and auxiliary tasks, but that avoids the minimum of the adversarial task.}
    \label{fig:multi-task-theory}
\end{figure}

Similarly to how learning happens in humans, multi-task architectures aim at simultaneously learning multiple tasks that are related to each other.
In principle, learning two related tasks generates mutually beneficial signals that lead to more general and robust representations than traditional or multimodal learning.
Figure~\ref{fig:multi-task-theory} a) and b) better illustrate the original concept proposed by \cite{caruana1997multitask}. 
Suppose that a complex model, e.g. a CNN, is trained on the main task M. In Figures~\ref{fig:multi-task-theory} (a) and (c) the optimization objective of M has two local minima, represented as the set \{\textit{a, b}\}. in Figure \ref{fig:multi-task-theory} (a), the auxiliary task A is related to the main task, with which it shares the local minimum in \textit{a}.
The joint optimization of M and A is thus likely to identify the shared local minima \textit{a} as the optimal solution \citep{caruana1997multitask}. 
The search is biased by the extra information given by task A towards representations that lay at the intersection of what could be learned individually for each task. 
In Figure~\ref{fig:multi-task-theory} (b), the auxiliary task is totally unrelated to the main task. 
No local minima are shared in this case and a negative transfer may happen without positive improvements to the performance.
%
Multi-task architectures divide into two families depending on the hard or soft sharing of the parameters.  
In architectures with hard parameter sharing such as the one proposed in this paper, multiple supervised tasks share the same input and some intermediate representation. 
The parameters learned up to this intermediate point are called \textit{generic parameters} since they are shared across all tasks. 
In soft parameter sharing, the weight updates are not shared among the tasks and the parameters are task-specific, introducing only a soft constraint on the training process \citep{duong2015low}. 

As explained by \cite{caruana1997multitask}, multi-task learning leads to various benefits if the tasks are linked by a valid relationship.
For instance, the generalization error bounds are improved and the risk of overfitting is reduced\citep{baxter1995learning}.
The speed of convergence is also increased since fewer training samples are required per task \citep{baxter2000model}. 
{Because of this,} multi-task learning has been successful in various applications, such as natural language processing \citep{subramanian2018learning}, computer vision \citep{kokkinos2017ubernet}, autonomous driving \citep{leang2020dynamic}, radiology \citep{vincent2021miccai} and histology \citep{gamper2020multi,marini2022multi}. 
The preliminary work by \cite{gamper2020multi}, in particular, shows a decrease in the loss variance as an effect of multi-task for oral cancer, suggesting that this work may have a high potential for histology applications.
Depending on the applications and on the loss functions used to represent the multiple tasks, multiple strategies exist for weighting each task contribution in the objective function \cite{gong2019comparison}.  
Alternatively to uniform weighting, for example, dynamical task re-weighting during training was proposed by \cite{leang2020dynamic} and \cite{kendall2018multi}. {The latter, in particular, exploits uncertainty estimates to weight each task, making the model convergence more robust to distribution shifts and unseen samples.} 

{The existing frameworks do not consider yet the combination of multitask learning with adversarial tasks. In Figure~\ref{fig:multi-task-theory} (c), we illustrate how our contribution in this paper differs from existing studies. For instance, we extend the multi-task framework to introduce an adversarial task C that is learned by adversarial training~\cite{ganin2016domain}.
In this case, the main task M has local minima in \{\textit{a, b, c}\}, but the minimum in \textit{c} is also a solution to the adversarial task C. We hypothesize that, by being adversarial to C, the optimization will be likely to prefer solutions that satisfy M and A, while avoiding solutions that satisfy the adversarial task C. 
This is because the main and auxiliary tasks (M and A) work in a cooperative setting with the feature encoder to learn a representation that will minimize both of their losses (Caruana, 1997). On the other hand, the adversarial task (i.e. C) plays an adversarial game with the feature encoder as it is described by Ganin et al. The encoder will optimize the representations so as to maximize the loss on the adversarial task, reducing the possibilities of recovering the information about the undesired target no matter how hard the adversarial branch tries.
The architecture in this work implements the extension of multitask learning to include adversarial tasks, and this is one of the main contributions of our work. 
}

%

%

\subsection{Adversarial Learning}
Proposed by \cite{ganin2016domain}, adversarial learning introduced a novel approach to solving the so-called problem of domain adaptation, namely the minimization of the domain shift in the distributions of the training (also called source distribution) and testing data (i.e. target).
Typically treated as either an instance re-weighting operation \citep{gong2013connecting} or as an alignment problem \citep{long2015learning}, domain adaptation is handled by adversarial learning as the optimization of a domain confusion loss. 
A domain classifier discriminates between the source and the target domains during training and its parameters are optimized to minimize the error when discriminating the domain labels. 
This can be extended to more than two domains by a multi-class domain classifier.
The adversarial learning of domain-related features is obtained by a gradient reversal operation on the branch learning to discriminate the domains. 
Because of this operation, the network parameters are optimized to maximize the loss of the domain classifier, thus making multiple domains impossible to distinguish one from another in the internal network representation. 
This causes competition between the main task and the domain branch during training which is referred to as a min-max optimization framework. 
As a downside, the optimization of adversarial losses may be complicated, with the min-max operation affecting the stability of the training \citep{ganin2016domain}. Convergence can be promoted, however, by activating and de-activating the gradient reversal branch according to a training schedule as in \cite{lafarge2017domain}.

{Adversarial learning is one building block of our architecture, since we incorporate the adversarial task as an additional branch of our multi-task architecture. Differently from the main work by \cite{ganin2016domain}, we introduce an additional hyper-parameter that controls the activation of the gradient reversal operator, so that the gradient flow is reversed for the adversarial tasks and kept unchanged for the rest of the auxiliary tasks in the architecture. }
\subsection{Concept-based Interpretable Modelling}
{One important feature of our architecture is that it introduces the learning of interpretable, high-level features as additional tasks to regularize the training process. This introduces transparency to the architecture, since the additional tasks introduce interpretable inductive biases during training. We revise, in the following, the concept of interpretability of deep learning, clarifying how this constitutes an important building block for our contributions.}
Interpretability of deep learning has become increasingly important over the past decade, which saw the definition of a plethora of methods and approaches with discording terminologies~\citep{graziani2022global}. 
In the context of digital pathology, interpretability analyses mainly focus on achieving post-hoc explanations rather than direct interpretable modelling~\citep{reviewXAIMI}.
Linear models, in particular, were proposed as an inherently interpretable approach to probe the internal activations of the network after training.
Regression Concept Vectors (RCVs) and Concept Activation Vectors (CAVs) were shown to generate insightful explanations in terms of diagnostic measures~\citep{kim2017interpretability,graziani2018regression,graziani2019improved,graziani2019interpreting,yeche2019ubs}. 
RCVs solve a linear regression problem in the space of the internal activations of a CNN layer. 
They were used to evaluate the relevance of a \textit{concept}, e.g. a clinical feature such as nuclei area or tumour extension, in the model output generation. 
The performance of the linear regressor indicated how well the concept was learned. 
Both CAVs and RCVs constitute a baseline of linear interpretability of CNNs, formalized for applications in the medical domain as concept attribution \citep{graziani2020concept}. 

No possibility is given by the existing interpretability methods to act on the training process and modify the learning of a concept. 
It is not possible, for example, to discourage the learning of a confounding concept, e.g. domain, staining, watermarks. 
Similarly, the learning of discriminant concepts cannot be further encouraged.
With this paper, we aim at addressing this gap. 
\section{Methods}
\label{sec:methods}
\subsection{Proposed Architecture}
The architecture for guiding the training of CNNs is described in this section for a general application with pre-defined features. This general framework can be applied to multiple tasks. The diagnosis of cancerous tissue in breast microscopy images is proposed in this paper as an application for which the implementation details are described in Sections~\ref{subsec:pathology-architecture} and \ref{subsec:extra-targets}.

In the following, we clarify the notation used to describe the model. 
We assume that a set of $N$ observations, i.e. the input images, is drawn from an unknown underlying distribution and split into a training subset $\{\bm{x}_i\}_{i=1}^n$ and a test subset $\{\bm{x}_i\}_{i=n+1}^N$.
The main task, namely the one for which we aim at improving the generalization, is the prediction of the image labels $\bm{y}=\{y_i\}_{i=1}^n$, for which ground truth annotations are available. 
A CNN of arbitrary structure is used as a feature encoder, of which the features are then passed through a stack of dense layers. 
The model parameters up to this point are defined as $\bm{\theta}_f$.
The parameters of the label prediction output layers are identified by $\bm{\theta}_y$.
The structure described up to this point replicates a standard CNN with a single main task branch that is addressing the classification. 
The remaining parameters of the architecture implement (i) the learning of auxiliary tasks by multi-task learning \citep{caruana1997multitask} and (ii) the adversarial learning of detrimental features to induce invariance in the representations, as in the domain adversarial approach by \cite{ganin2016domain}.
We combine these two approaches by introducing $K$ extra targets representing desired and undesired tasks that must be introduced to the learning of the representations. The targets are modeled as the prediction of the feature values $\{c_{k,i}\}_{i=1}^N$, where $k\in{1,\dots,K}$ is an index representing the extra task being considered. The feature values may be either continuous or categorical.  
Additional parameters $\bm{\theta}_{k}$ are trained in parallel to $\bm{\theta_y}$ for the $K$ extra targets.
We refer to all model outputs for all inputs $\bm{x}$ as $f(\bm{x}) \in \mathcal{R}^{K+1}$ .
%
\begin{figure}[t!]
    \centering
    \includegraphics[width=0.95\textwidth]{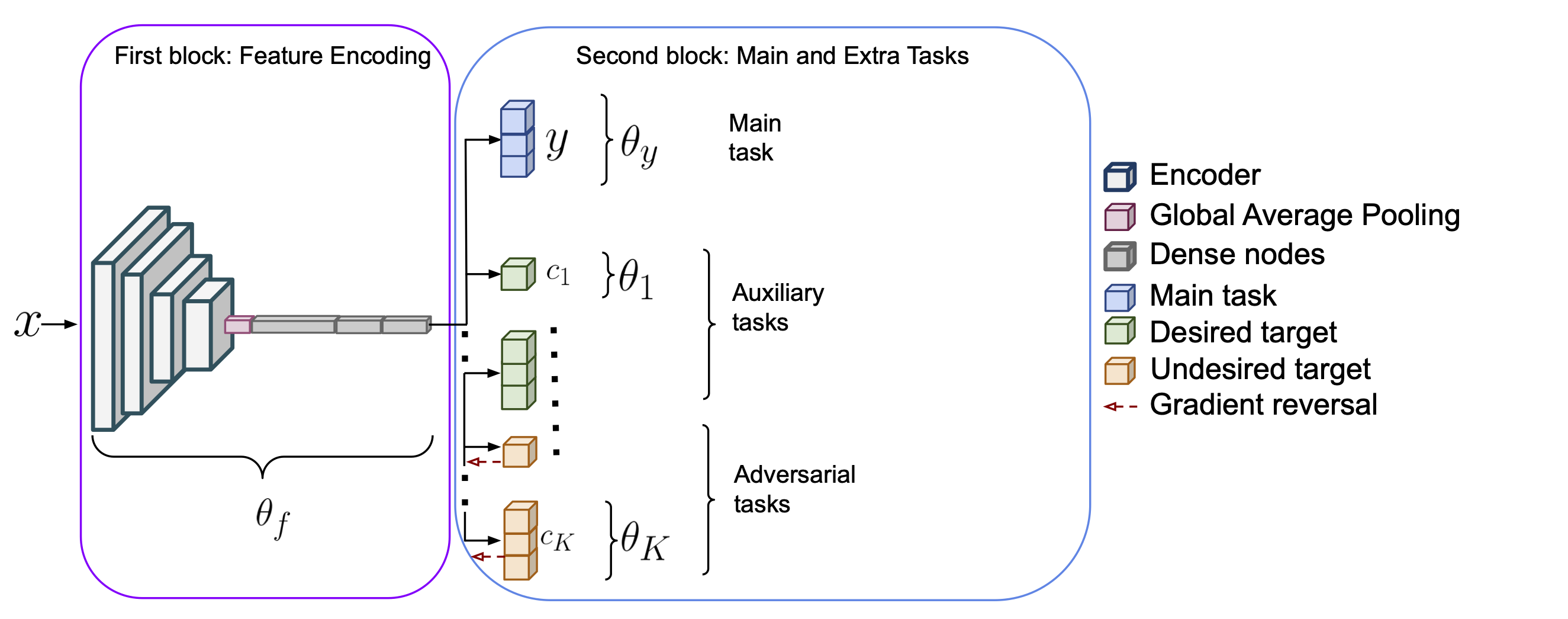}
    \caption{Multi-task adversarial architecture}
    \label{fig:architecture}
\end{figure}

The architecture is illustrated in Figure~\ref{fig:architecture} and consists of two blocks. The first block is used to extract features from the input images. A state-of-the-art CNN of arbitrary choice without the decision layer is used as a feature encoder generating a set of feature maps.
The feature maps are passed through a Global Average Pooling (GAP) operation that is performed to spatially aggregate the responses and connect them to a stack of dense layers. For this specific architecture, we use a stack of three dense layers of 1024, 512, and 256 nodes respectively. 
The second block comprises one branch per task, taking as input the output of the first block. The main task branch consists of the prediction of the labels $\mathbf{y}$ and has as many dense nodes as there are of unique classes in $\mathbf{y}$. For binary classification tasks, e.g. discrimination of tumorous against non-tumorous inputs, the main task branch has a single node with a sigmoid activation function. $K$ branches are added to model the extra targets.
We refer to \textit{extra} tasks for all the additional targets to the main task whether desired or undesired. 
\textit{Auxiliary} tasks refer to the modeling of the desired targets, while \textit{adversarial} tasks refer to that of undesired targets. 
The extra tasks are modeled by linear models as in \cite{graziani2018regression}. For continuous-valued targets, the extra branch consists of a single node with a linear activation function. For categorical targets, the extra branch has multiple nodes followed by a softmax activation function. 
A gradient reversal operation \citep{ganin2016domain} is performed on the branches of the undesired targets to discourage the learning of these features. 
\subsection{Objective Function}
The objective function of the proposed architecture balances the losses of the main task and the extra tasks for the desired and undesired targets. This is obtained by a combination of multi-task and adversarial learning. 
The main task loss is $\mathcal{L}_y^i(\bm{\theta_f},\bm{\theta_y})=\mathcal{L}_y(\bm{x}_i,y_i;\bm{\theta_f},\bm{\theta_y})$, where $\bm{\theta}_f$ are the parameters of the first block (namely of the CNN encoder and the dense layers) in Figure~\ref{fig:architecture} and $\bm{\theta}_y$ those of the main task branch in the second block of the same figure. 
The extra parameters $\bm{\theta}_{k}$ ($k\in{1,\dots,K}$) are trained for the branches of the desired and undesired target predictions, with the loss being $\mathcal{L}_{k}^i(\bm{\theta}_f, \bm{\theta}_{k}) = \mathcal{L}_{k}(\bm{x}_i,c_{k,i};\bm{\theta}_f,\bm{\theta}_{k})$.


Training the model on $n$ training and $(N-n)$ testing samples consists of optimizing the function: 
\begin{equation}\label{eq:optimization_function}
    E(\bm{\theta}_y, \bm{\theta}_f, \bm{\theta}_{1}, \dots, \bm{\theta}_{K}) = \lambda_{m} \frac{1}{n} \sum_{i=1}^n \mathcal{L}_y^i(\bm{\theta}_f,\bm{\theta}_y) +  \sum_{k=1}^K \lambda_k \frac{1}{N} \sum_{i=1}^N \mathcal{L}_{k}^i(\bm{\theta}_f,\bm{\theta}_{k}).
\end{equation}{}

The gradient update is:
\begin{equation}\label{update_f}
    \bm{\theta}_f \leftarrow \bm{\theta}_f - \left( \lambda_m  \frac{\partial\mathcal{L}_y^i}{\partial\bm{\theta}_f}+ \sum_{k=1}^K \lambda_k \alpha_k  \frac{\partial\mathcal{L}_{k}^i}{\partial\bm{\theta}_f} \right),
\end{equation}
\begin{equation}\label{update_y}
    \bm{\theta}_y \leftarrow \bm{\theta}_y - \lambda_m\frac{\partial\mathcal{L}_y^i}{\partial\bm{\theta}_y},
\end{equation}
\begin{equation}\label{update_c}
    \bm{\theta}_{k} \leftarrow \bm{\theta}_{k} -
    \lambda_k  \frac{\partial\mathcal{L}_{k}^i}{\partial\bm{\theta}_{k}},
\end{equation}
where $\lambda_m$ and $\lambda_k$ are positive scalar hyperparameters to tune the trade-off between the losses. For each extra branch, the hyperparameter $\alpha_k \in \{-1, 1\}$ is used to specify whether the update is adversarial or not. A value of $\alpha_k = -1$ activates the gradient reversal operation and starts an adversarial competition between the feature extraction and the corresponding $k$\textsuperscript{th} extra branch. 
The main task is only trained on the training data, since $\mathcal{L}_y^i=0$ for $i>n$ in Eq.~\eqref{update_f} and \eqref{update_y}. 
Following the work in~\cite{ganin2016domain}, the additional branches can be trained on slightly different dataset splits. If additional task labels are available for the test set, they can be included in the training of the additional branches. {Moreover, if only partially labeled data are available, they can also be introduced in the training of the additional branches. For instance, the gradient updates can be kept only for the labeled portion of the data, setting the loss to zero for the unlabeled inputs.}
\subsection{Loss weighting strategy}
The proposed architecture requires the combination of multiple objectives in the same loss function. The vanilla formulation in Eq.~\ref{eq:optimization_function} simply performs a weighted linear sum of the losses for each task. 
This is the predominant approach used in prior work with multi-objective losses \citep{gong2019comparison} and adversarial updates \citep{ganin2016domain,lafarge2017domain}.
The appropriate choice of weighting of the different task losses is a major challenge of this setting. The tuning of the hyperparameters may reveal tedious and non-trivial due to the combination of classification and regression tasks with different ranges of the loss function values (e.g. combining the bounded binary cross-entropy loss in [0,1] with the unbounded mean squared error loss).

An optimal weighting approach may be learned simultaneously with the other tasks by adding network parameters for the loss weights $\lambda_m$ and $\lambda_k$. 
The direct learning of $\lambda_m$ and $\lambda_k$ would just result in weight values quickly converging to zero. 
\cite{kendall2018multi} proposed a Bayesian approach that makes use of the homoscedastic uncertainty of each task to learn the optimal weighting combination. 
In loose words, homoscedastic uncertainty reflects a task-dependent confidence in the prediction. 
The main assumption to obtain an uncertainty-based loss weighting strategy is that the likelihood of the task output can be modeled as a Gaussian distribution with the mean given by the model output and a scalar observation noise $\sigma$:
\begin{equation}
    p(\bm{y}|f(\bm{x})) = \mathcal{N}(f(\bm{x}), \sigma^2)
\end{equation}
This assumption is also applied to the outputs of the additional tasks. 
The loss weights $\lambda_m$ and $\lambda_k$ are then learned by optimizing the minimization objective given by the negative log likelihood of the joint probability of the task outputs given the model predictions. 
To clarify this concept, let us focus on a simplified architecture with the main task being the logistic regression of binary labels (e.g. tumor v.s. non-tumor) with noise $\sigma_1$ and one auxiliary task consisting of the linear regression of feature values $\bm{c}=\{c_i\}_{i=1}^N$, with noise $\sigma_2$. 
The minimization objective for this multi-task model is:
\begin{equation}
\label{eq:uncertainty}
    -\log p(\bm{y},\bm{c}|\bm{f}(\bm{x})) \propto \frac{1}{2\sigma_1^2} \mathcal{L}_y(\bm{\theta}_f, \bm{\theta}_y) + \frac{1}{2\sigma_1^2} \mathcal{L}_k(\bm{\theta}_f, \bm{\theta}_k) + \log \sigma_1 + \log \sigma_2
\end{equation}

By minimizing Eq.~\ref{eq:uncertainty} with respect to $\sigma_1$ and $\sigma_2$, the optimal weighting combination is learned adaptively based on the data \citep{kendall2018multi}. As $\sigma_1$ increases, the weight for its corresponding loss decreases, and vice-versa. The last term $\log \sigma_1 + \log \sigma_2$, besides, acts as a regularizer discouraging each noise to increase unreasonably.
This construction can be extended easily to multiple regression outputs and the derivation for classification outputs is given in 
\cite{kendall2018multi}. 
%
%
%
\subsection{Dataset}
\begin{table}[b]
\fontsize{9}{10}\selectfont
    \centering
    \caption{Summary of the train, validation and internal test splits.}
    \label{tab:data-splits}
\begin{tabular}{cc|c|ccccc|cc}
& & Cam16 & \multicolumn{5}{c}{ Cam17 (5 Centers)}\vline & \multicolumn{2}{c}{ PanNuke (2 Folds) } \\
& Label & & C. 0 & C. 1 & C. 2 & C. 3 & C. 4 & F. 1 & F. 2  \\\hline
\multirow{2}{*}{ Train } & Neg. & 12954 & 31108 & 25137 & 38962 & 25698 & 0 & 1425 & 1490  \\
& Pos. & 6036 & 8036 & 5998 & 2982 & 1496 & 0 & 2710 & 2255 \\\hline
\multirow{2}{*}{ Val. } & Neg. & 0 & 325 & 0 & 495 & 0 & 0 & 0 & 0  \\
& Pos. & 0 & 500 & 0 & 500 & 0 & 0 & 0 & 0 \\\hline
\multirow{2}{*}{ Int. Test } & Neg. & 0 & 0 & 274 & 483 & 458 & 0 & 0 & 0  \\
& Pos. & 0 & 500 & 999 & 0 & 0 & 0 & 0 & 0 \\
\end{tabular}
\end{table}

\begin{table}[h!]
\fontsize{9}{10}\selectfont
    \centering
    \caption{External test splits}
    \label{tab:test-splits}
    \begin{tabular}{cc|c|ccccc|ccc}
    &Label & Cam17 C. 4 & {PanNuke Fold3} \\\hline
       \multirow{2}{*}{ CamExt} & Neg. & 500 & 0  \\
& Pos.  & 500 & 0 \\\hline
           \multirow{2}{*}{ PanExt } & Neg. & 0 & {} 480 \\
& Pos. & 0 & {} 395 \\
    \end{tabular}
\end{table}
The experiments are run on three publicly available datasets, namely Camelyon 16, Camelyon 17~\citep{litjens20181399} and the breast subset of PanNuke~\citep{gamper2019pannuke,gamper2020multi} {(The data are available for download at the links \url{https://camelyon17.grand-challenge.org} and \url{https://warwick.ac.uk/fac/cross\_fac/tia/data/pannuke}.} 
The Camelyon challenge collections of 2016 and 2017 contain respectively 270 and 899 WSIs. 
All training slides of both challenges contain annotations of metastasis type (i.e. negative, macro-metastases, micro-metastases, isolated tumor cells), and 320 images contain manual segmentations of tumor regions. 
The analysis also includes the breast tissue scans of the PanNuke dataset, for which multiple nuclei types were annotated by the semi-automatic instance segmentation tool described in~\cite{gamper2019pannuke}. Labels of neoplastic, inflammatory, connective, epithelial and dead nuclei are given together with the images by the dataset creators. 
Training, validation and test splits are built on these two datasets as reported in Table~\ref{tab:data-splits}. The pre-existing PanNuke folds are used, two of which (i.e. fold 1 and fold 2) are used in the training set.
{ For external validation of our model we create the two splits described in Table~\ref{tab:test-splits}, namely the CamExt and the PanExt splits. CamExt is built by leaving out from training the images of Camelyon17 that were acquired at one specific acquisition center, namely center 4 in the original collection. 
The lesions of patients in this center have a much larger incidence of macro lesions than the patients in the internal validation set, with 26 slides out of 100 reporting a tumor diameter larger than 2.0 mm. Because of the expanded tumor size, true positives may be relatively easy to obtain on this set since the patches are less likely to be sampled from the tumor borders, and thus do not contain normal and tumorous tissue at the same time.
The PanExt dataset is further used to test model performance, and it comprises the data in the testing split of the PanNuke collection, namely the images in Fold 3 of the original dataset.
We extract 775
image patches of which 480
contain tumor and 295
represent healthy tissue without tumorous cells. 
The target labels for the additional tasks were not computed for this dataset, hence these images were never seen by the model during training at the patch, slide and patient levels.
}
All WSIs are pre-processed in the same way. Patches of $224 \times 224$ pixels are extracted at the highest magnification level and the staining variability is reduced by Reinhard normalization \cite{reinhard2001color,staintools}.  
{ The PanNuke images do not depict an entire WSI but only a small portion, from which we also extract image patches of $224 \times 224$ pixels.}
{ During training we perform oversampling with slightly overlapping patches by extracting patches located in the center, upper left, upper right, bottom left and bottom right corners of the image. This fully covers the WSI portion depicted in the original PanNuke image content and it is applied to balance the domain under-representation of these input images.}

%
\begin{figure}
    \centering
    \includegraphics[width=\textwidth]{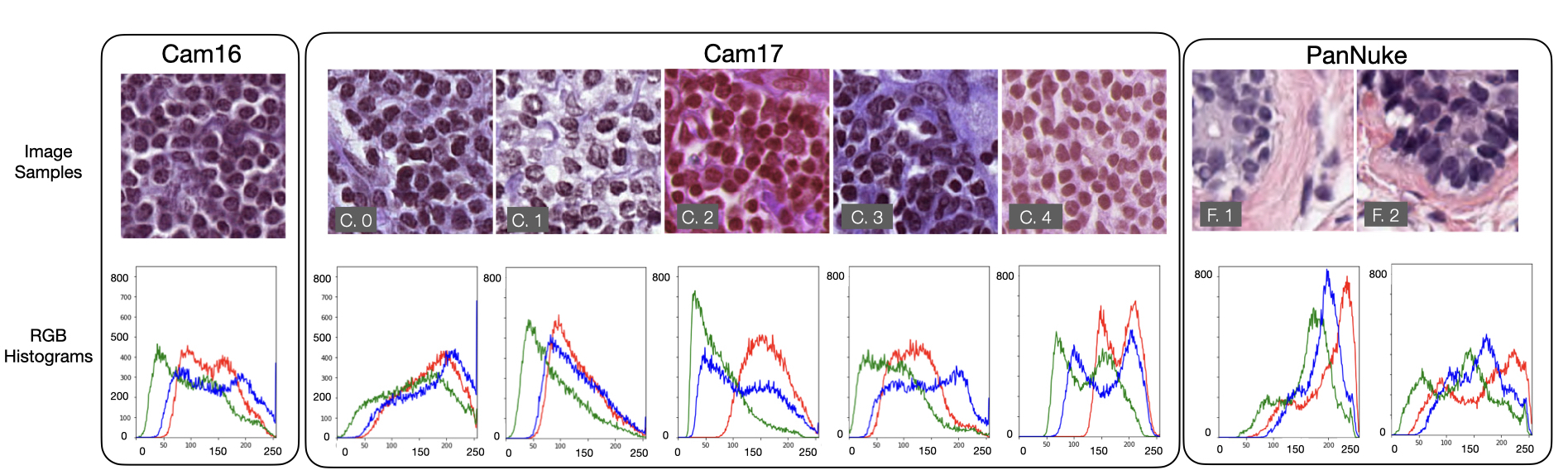}
    \caption{An example of the visual heterogeneity in the images due to shifts in the acquisition protocols of each collection center. In the top row, we show image samples from healthy tissue for each center. In the bottom row, the respective RGB (Red, Green and Blue channels) histograms of the images. }
    \label{fig:shift}
\end{figure}
\subsection{Main task and architecture backbone}
\label{subsec:pathology-architecture}
The main task that we address is the binary classification of input images that include tumor tissue from those without tumor. 
Inception V3 pretrained on ImageNet~\citep{szegedy2016rethinking} is used as the backbone CNN for feature encoding. 
{Following the observations in\citep{graziani2019visualizing},}
the parameters up to the last convolutional layer are kept frozen to avoid overfitting to the pathology images\footnote{{ A total of 7,804,427 trainable parameters are used in the first block for the feature extraction with InceptionV3 as the convolutional backbone and 6,427,392 with ResNet50. In the second block there are 257 to 2,827 trainable parameters, depending on the number of tasks being trained jointly.
}}.
The output of the CNN is passed through the GAP and the three fully connected layers as illustrated in Figure~\ref{fig:architecture}. 
The fully connected layers have respectively 2048, 512 and 256 units.
A dropout probability of 0.8 and L2 regularization are added to these three fully connected layers to avoid overfitting. 
{This configuration was obtained by optimizing performance only on the main task and by searching the hyperparameter space to identify the optimal dense block width, regularization strength and dropout probability.}
The main task is the detection of patches containing tumor as a binary classification task. The branch consists of a single node with sigmoid activation function connected to the output of the third dense layer. 
The architecture as described up to here, hence without extra branches, is used as the baseline for the experiments. 
The additional tasks consist of either the linear regression or the linear classification of continuous or categorical labels respectively. 
For linear regression, the additional branch is a single node with linear activation function. The Mean Squared Error (MSE) between the predicted value and the label is added to the optimization function in Eq. \ref{eq:optimization_function}.
For the linear classification, the extra branch has a number of dense nodes equal to the number of classes to predict and a softmax activation function, also connected to the third dense layer. The Categorical Cross-Entropy (CCE) loss is added to the optimization in Eq.~\ref{eq:optimization_function}. 
Further details about the extra branches used for the experiments are given in Section~\ref{subsec:extra-targets}. 
 
The architecture is trained end-to-end with mini-batch Stochastic Gradient Descent (SGD) with standard parameters (learning rate of $10^{-4}$ and Nesterov momentum of 0.9). 
{The main task is learned by optimizing the class-Weighted Binary Cross Entropy (WBCE) loss, where positive samples are given a higher weight than the negative ones to counter-balance their under-representation in the training set. The weights are set so that they sum to one. The weight for the positive class corresponds to the number of samples missing to reach one, namely one minus the ratio of positive samples, i.e. 0.82. Similarly, the weight for the negative class is set to 0.18.}

We evaluate the convergence of the network by early stopping on the total validation loss with patience of 5 epochs. 
The Area Under the ROC Curve (AUC) is used to evaluate model performance. 
For each experiment, we perform five runs with multiple initialization seeds to evaluate the performance variation due to initialization. 
The splits are kept unchanged for the multiple seed variations. 
To evaluate the performance on multiple test splits, we perform bootstrapping of the test sets. 
A number of 50 test sets of 7589 images (the total number of test images in the two sets) are obtained by sampling with replacement from the total pool of testing images. 
This method evaluates the variance of the test set without prior assumption on the data distribution and it shows the performance difference due to variation of the sampling of the population. 



\subsection{Configuration of the additional targets}
\label{subsec:extra-targets}
\begin{figure}
    \centering
    \includegraphics[width=1.\linewidth]{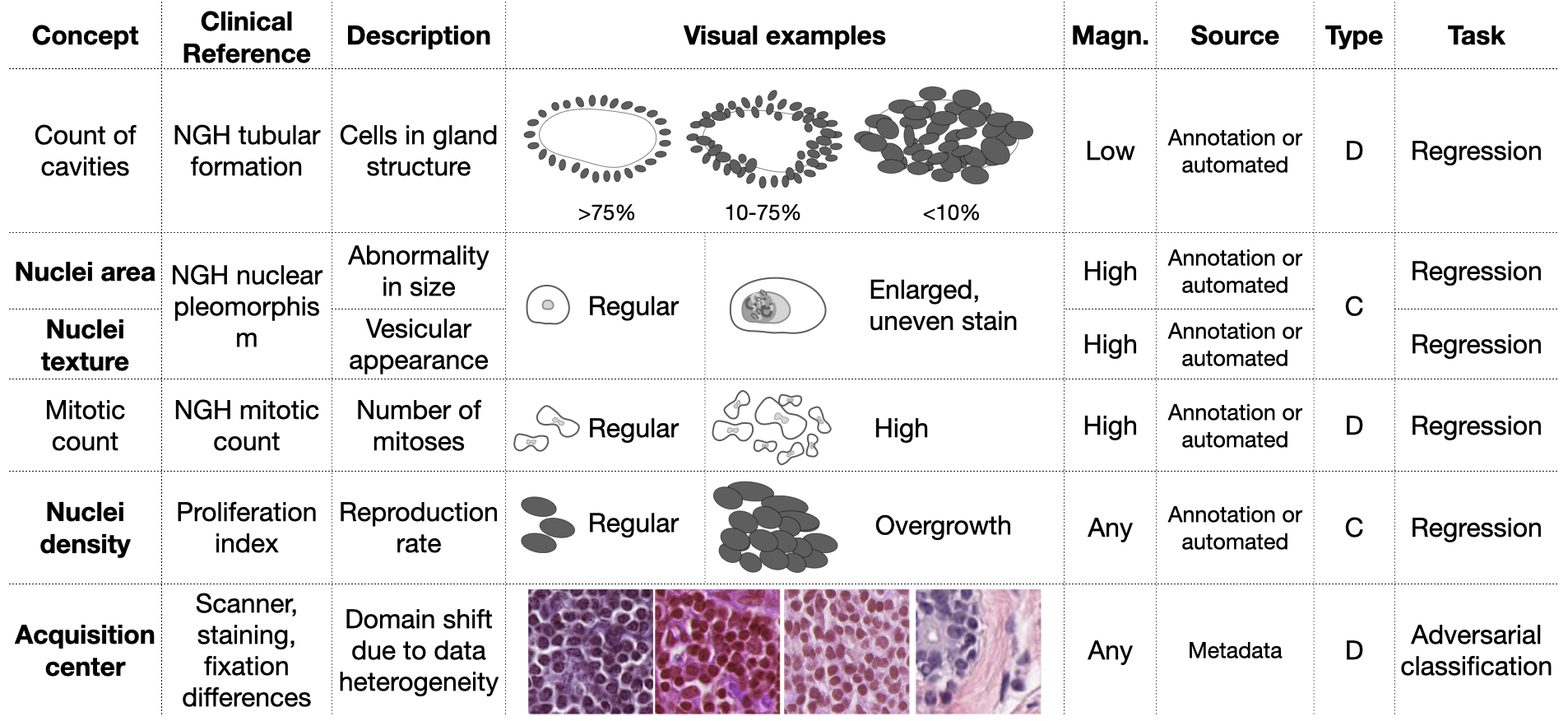}
    \caption{Control targets for breast cancer. C and D stand for continuous and discrete respectively. {The targets used in this work are highlighted in bold.}}
    \label{fig:extra-targets}
\end{figure}

The experiments focus on the integration of four desired and one undesired targets with multiple combinations. 
The auxiliary targets relate to the main task, being important diagnostic features. We expect that learning of these desired features will improve the solution robustness and generalization of the model. Discarding the undesired targets may improve the invariance of the learned features to confounding factors. 
The Nottingham Histologic Grading (NHG) of breast tissue identifies the key diagnostic features for breast cancer \citep{bloom1957histological}. By analyzing this we derived the desired and undesired features that are illustrated in Figure~\ref{fig:extra-targets}. 
From this set, we retain cancer indicators at the nuclear level, since the input images are at the highest magnification. 
We model the variations of the nuclei size, appearance (e.g. irregular, heterogeneous texture) and density shown in Figure~\ref{fig:extra-targets} as real-valued variables. 
{ Because of the heterogeneity of the data, we also guide the network training to discard information about the image acquisition center, which is modeled as an undesired target. The staining variability constitutes part of the shift, as differences are visible by the naked eye as shown in Figure~\ref{fig:shift}. The staining variability, however, is only one of the components that contribute to creating the distributional shift among centers. Within the domain shift, other sources of variability further enhance the heterogeneity of the data, e.g. tissue fixation times, processing temperature and scanner resolution. These factors vary across institutions and jointly contribute to the distributional shifts observed among centers\footnote{For this reason, the staining normalization during preprocessing does not interfere with detecting the acquisition site.}. }
{}

Hand-crafted features representing the variations in the nuclei size and appearance are automatically extracted either from the images of from the nuclear contours.
The nuclear contours are available in the form of manual annotations only for the PanNuke data. 
{The manual delineation of nuclei contours may be cumbersome, costly and not entirely reflect the standard clinical routine procedure for cancer diagnosis. For this reason, we include  nuclei segmentation labels that are obtained by an automatic segmentation model.}
For the images in Camelyon, automated contours of the nuclei are obtained by the multi-instance deep segmentation model in \cite{otaloraasystematic}.
This model is a Mask R-CNN model~\citep{He_2017_ICCV}, fine-tuned from ImageNet weights on the Kumar dataset for the nuclei segmentation task~\citep{kumar}. 
The R-CNN identifies nuclei entities {in the individual patches obtained from each WSI}, and it then generates pixel-level masks by optimizing the Dice score. 
{ From the pre-existing labeled dataset of 30 annotated WSIs, we generate weak labels for over a thousands slides in the Camelyon datasets.}
ResNet50~\citep{He_2017_ICCV} is used for the convolutional backbone as in~\citep{otaloraasystematic}. 
The network is optimized by SGD with standard parameters (learning rate of 0.001 and momentum of 0.9). 

\begin{figure}
    \centering
     \begin{subfigure}{0.24\linewidth}
        \includegraphics[width=\linewidth]{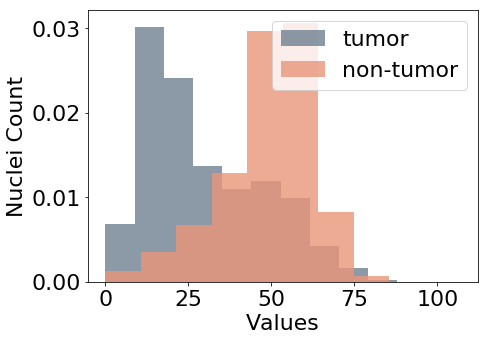}
        \caption{}
        \label{fig:ncountdist}
    \end{subfigure}
    \hfill
    \begin{subfigure}{0.24\linewidth}
        \includegraphics[width=\linewidth]{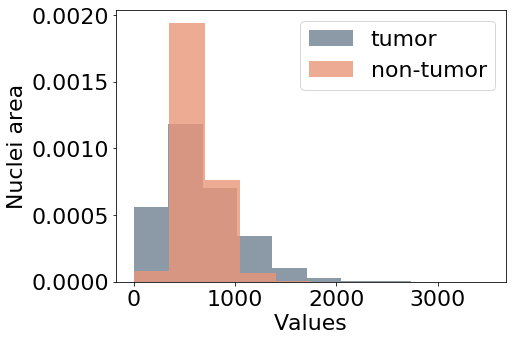}
        \caption{}
        \label{fig:nareadist}
    \end{subfigure}
    \begin{subfigure}{0.24\linewidth}
        \includegraphics[width=\linewidth]{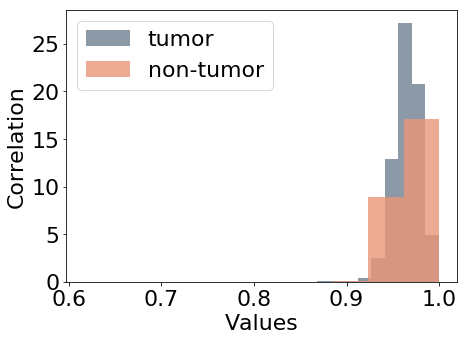}
        \caption{}
        \label{fig:corrdist}
    \end{subfigure}
\begin{subfigure}{0.24\linewidth}
        \includegraphics[width=\linewidth]{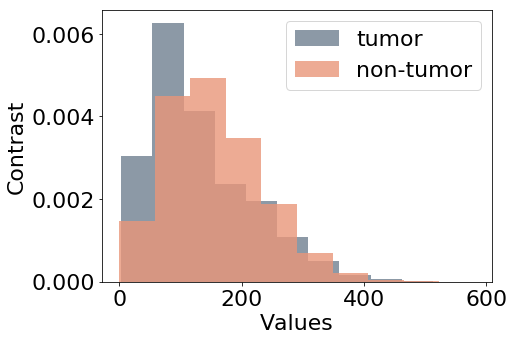}
        \caption{}
        \label{fig:contrdist}
    \end{subfigure}
    \caption{{Distribution of nuclei count (a) nuclei area (b), correlation (c) and contrast (d) values in the training data. Nuclei count is bimodal for the tumor and non-tumor classes.}}
    \label{fig:targets.distribution}
\end{figure}

{Figure~\ref{fig:targets.distribution} shows the distribution of the feature values considered as additional targets for this study.}
Nuclei \textit{density} in Figure~\ref{fig:ncountdist} is estimated by counting the nuclei in each {patch.}
The number of pixels inside nuclear contours is averaged for each input {patch} to represent variations of the nuclei area (b), referred to as \textit{area} in the experiments.

Haralick descriptors of texture {correlation} and {contrast} \citep{haralick1979statistical} are also extracted from the {patches} as in \cite{graziani2018regression}, {for which the value distributions are shown in Figures~\ref{fig:corrdist} and~\ref{fig:contrdist} respectively}. 
Being continuous and unbounded measures, the values for these features are normalized to have zero mean and unitary standard deviation before training the model. 
In the paper, we refer to these features as \textit{area}, \textit{density}, \textit{contrast} and \textit{correlation}. 
%
The values of these features are used as prediction labels for the auxiliary target branches, that are also named as the feature that they should predict. 
These auxiliary branches perform a linear regression task, trying to minimize the Mean Squared Error between the predicted value of the feature and the extracted values used as labels. 

\begin{figure}
    \centering
    \includegraphics[width=\linewidth]{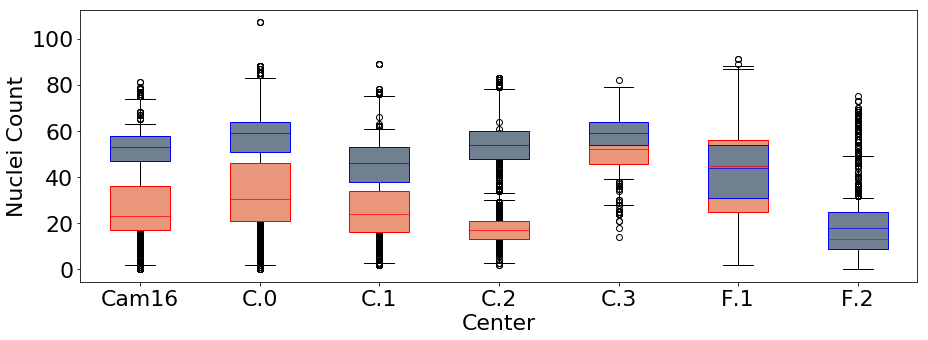}
    \caption{ Boxplots of the nuclei count values in the training dataset, separated for each center of the adversarial task and for the tumor and non-tumor classes of the main task. }
    \label{fig:dist.breakdown}
\end{figure}

{The center that performed the data acquisition may act as a confounder on both the main task and the additional, desired targets. }{Figure~\ref{fig:dist.breakdown}, for example, illustrates how the acquisition center impacts the nuclei density, i.e. the observed values of nuclei count.} For this reason, we use information about the center that performed the data acquisition to model an adversarial task, using the center value that is present in the dataset as metadata. We model it as a categorical variable that may take values from 0 to 7, namely one for each known center in the data. 

Since there is no specific information on acquisition centers in Camelyon16 and PanNuke, these have been modeled as two distinct acquisition centers in addition to the five known centers of Camelyon17. This information is partly inaccurate, since we know that in both datasets more than a single acquisition center was involved \citep{camelyon,gamper2020pannuke}. The noise introduced by this information may limit the benefits introduced by the adversarial branch but it should not affect the performance negatively. In the future, unsupervised domain alignment methods may also be  explored. The prediction of this variable is added to the architecture as an undesired target branch, referred to as \textit{center} in the experiments.
\section{Results}
\label{sec:results}
\subsection{Main task predictivity from the auxiliary targets}
{We validate the predictive power of the features that we selected as auxiliary targets by a simple test. We train multi-layer perceptron models (MLP) with 20 hidden layers to predict the main task from the auxiliary label values that are, in this case, passed as input features and not as additional, multi-task targets.
The models were trained with BCE loss and with the Adam optimizer with standard parameters such as learning rate of 0.01, and early stopping patience of 15. 
We evaluate the predictive AUC on the internal test set, which is reported in Table~\ref{tab:lastreviewres}. 
As the results show, the main task predictions are better than random guessing for all the models that we trained, suggesting that the auxiliary labels contain information about the main task. Using multiple features at the same time (as in model ID MLP4), moreover, improves the performance of the classifier, showing that each auxiliary target contributes to the discrimination of the classes in the main task. 
}

    
\begin{table}[!htbp]
\fontsize{7}{11}\selectfont
    \centering
    \caption{Main task test AUC of a MLP trained on the auxiliary labels. The better than chance performance on the internal testing set shows that the labels contain information that can be beneficial for the learning of the main task.}
    \label{tab:lastreviewres}
    \begin{tabular}{c|c|c|c|c|c|c|c}
    \hline
Model &ID&random & area & count & contrast & \multicolumn{2}{c}{int. test}  \\\hline
MLP& MLP1 & \checkmark & & & & \multicolumn{2}{c}{$0.500$} \\\hline
MLP& MLP2 &  & & \checkmark& & \multicolumn{2}{c}{$0.630$} \\\hline
MLP& MLP3 &  & & \checkmark& \checkmark & \multicolumn{2}{c}{$0.660$} \\\hline
MLP& MLP4 &  & \checkmark & \checkmark& \checkmark & \multicolumn{2}{c}{$\textbf{0.684}$} \\\hline
\end{tabular} 
\end{table}

\subsection{Single-task baseline model}
In the baseline model, only the main task branch is trained and no extra tasks are used. 
{The results for the baseline models are shown in the first and second rows of Table~\ref{tab:results}, that are identified by unique IDs, namely model-ID 1 for the model with Inception V3 and model-ID R1 for the Res-Net model}. The experiments are performed on the dataset configurations in Tables~\ref{tab:data-splits} and~\ref{tab:test-splits}, where internal and external test sets are used to evaluate model generalization. Two columns are used to report the results on the internal (int.) and external (Cam.Ext.) test sets. 
{Table~\ref{tab:panext} reports the results on PanExt.}
Where not stated otherwise, the average AUC (avg. AUC) over ten repetitions with multiple initialization seeds is used for the evaluation. 
\subsection{Double-task combinations}
An ablation study is performed by adding a single additional task at a time to the baseline model. This study aims at identifying the benefits of encouraging each task individually. 
The desired and undesired targets described in Section~\ref{subsec:extra-targets} are added as additional branches to the architecture detailed in Section~\ref{subsec:pathology-architecture}.
The gradient reversal operation is only active for the \textit{center} branch.
The losses of each task are combined by two strategies, namely a vanilla and the uncertainty-based approach in~\cite{kendall2018multi}. 
In the vanilla configuration, the loss weight values are set to 1 for all branches.
The results for single task combinations are reported in Table~\ref{tab:results} with unique IDs ranging from 2 to 5. 
Model-ID 2, for example, is given by the combination of the main task branch with the additional task \textit{area}, namely of predicting the area of the nuclei in the images.
A single auxiliary branch already outperforms the baseline (internal avg AUC $0.819 \pm 0.001$, external avg. AUC $0,868 \pm 0.005$, {int. avg. F1 $0.783 \pm 0.002$, ext. avg. F1 $0.315 \pm 0.001$}), 
as for example in model-ID 3 by encouraging nuclei \textit{count} (internal avg AUC $0.836 \pm 0.005$, external avg. AUC $0,890 \pm 0.009$, {internal avg. F1 $0.768 \pm 0.001$, external avg. F1 $0.660 \pm 0.003$}) 
{The avg. F1 score of model-ID3 on the two test sets is at $0.746 \pm 0.009$, a significantly higher value than the $0.701 \pm 0.001$ of the baseline model}. 
On the external test set, the best generalization is achieved by adding \textit{count} as a desired target (ext. avg. AUC $0.890 \pm 0.009$). 
\subsection{Multi-task adversarial combinations}
The most promising branches are then combined to further improve performance.
The following combinations of multiple auxiliary and adversarial tasks are tested in the experiments: \textit{center} + \textit{density}, \textit{center} + \textit{area}, \textit{center} + \textit{density} + \textit{area}. 
The combination of all the branches leads to the best performance on the internal test { for both the models with the Inception V3 and ResNet backbones, namely in model-ID 8 and R8}. { These models yield} an increase from the baseline of 0.055 AUC points for model-ID 8 and 0.091 AUC points for R8.
{The highest performance on the internal test set is given by the R8 model with an AUC of $0.893 \pm 0.001$}
Model-ID 6 reports comparable performance on the external test set to model-ID 3. 
The addition of the \textit{center} adversarial branch in model-ID 6 leads to the best Inception V3 based model overall with average AUC on both internal and external sets at $0.824 \pm 0.006$  {and avg. F1 score $0.755 \pm 0.005$} for the uncertainty trained model.
This represents a significant improvement compared to the overall average AUC $0.79 \pm 0.001$ { and avg. F1 score $0.701 \pm 0.004$} of the baseline model, with $p-value<0.001$. The statistical significance of the results is evaluated by the non-parametric Wilcoxon test (two-sided) applied on the bootstrapping of the test set as described in Sec.~\ref{subsec:pathology-architecture}. 
{The performance on PanExt is reported for the baselines and the best performing models in Table~\ref{tab:panext}.}

\begin{table}[!htbp]
\fontsize{7}{11}\selectfont
    \centering
    \caption{Average AUC on the main task and standard deviations from different starting points of the network parameter initialization. Results for the vanilla and uncertainty based (unc.) weighting strategies. Inception V3 (IV3) and ResNet 50 (ResNet) are used as backbones. The adversarial task, i.e. \textit{center}, is marked by an overline.}
    \label{tab:results}
    \begin{tabular}{c|c|c|c|c|c|c|cc|cc}
    \hline
Model &ID&main & area & count & contrast & $\overline{center}$ & \multicolumn{2}{c}{int. test}\vline & \multicolumn{2}{c}{CamExt} \\\hline
IV3&1 & \checkmark & & & & & \multicolumn{2}{c}{$0.819 {\pm 0.001}$}\vline & \multicolumn{2}{c}{ $0.868 {\pm 0.005}$ } \\\hline
{ResNet} & {R1} & \checkmark & & & & & \multicolumn{2}{c}{$0.802 {\pm 0.003}$}\vline & \multicolumn{2}{c}{ $0.821 {\pm 0.004}$ } \\\hline
&&& & & & & vanilla & unc. & vanilla & unc. \\\hline
\multirow{7}{*}{IV3}&2&\checkmark & \checkmark & & & & $0.718{\pm 0.11} $ & $0.834 {\pm 0.01}$ & $ 0.560 {\pm 0.06}$ & $0.871 {\pm 0.01}$ \\
&3&\checkmark & & \checkmark & & & $0.853 {\pm 0.03} $ & $0.836 {\pm 0.005}$ & $ 0.874 {\pm 0.02}$ & $\textbf{0.890} {\pm \textbf{0.009}}$ \\
&4&\checkmark & & & \checkmark & & $0.854 {\pm 0.07}$ & $0.835 {\pm 0.008}$ & $ 0.883 {\pm 0.02}$ & $0.876 {\pm 0.007}$ \\
&5&\checkmark & & & & \checkmark & $0.845 {\pm 0.10}$ & $0.822 {\pm 0.005}$ & $ 0.884 {\pm 0.04}$ & $0.871 {\pm 0.005}$ \\
&6&\checkmark & & \checkmark & & \checkmark & $0.863 {\pm 0.06}$ & $0.841 {\pm 0.004}$ & $ 0.623 {\pm 0.10 }$ & $\
\textbf{{0.890}} {\pm \textbf{0.01}}$ \\
&7&\checkmark & \checkmark & \checkmark & & \checkmark & $0.838 {\pm 0.05}$ & $0.848 {\pm 0.003}$ & $ 0.490 {\pm 0.03}$ & $0.864 {\pm 0.01}$ \\
&8&\checkmark & \checkmark & \checkmark & \checkmark & \checkmark & $0.858 {\pm 0.02}$ & ${0.874} {\pm{ 0.009}}$ & $ 0.686 {\pm 0.20}$ & $0.825 {\pm 0,01}$ \\\hline
 {ResNet}& {R8}&\checkmark & \checkmark & \checkmark & \checkmark & \checkmark & n.a. &  {$\textbf{0.893}{\pm \textbf{0.001}}$} & n.a. & {${0.861} {\pm{ 0.01}}$} \\
\end{tabular} 
\end{table}
\begin{table}[!htbp]
\fontsize{8}{11}\selectfont
    \centering
    \caption{Performance on the PanExt dataset measured as the average AUC on the main task and standard deviations from different starting points of the network parameter initialization. The results are for the uncertainty-based weighting strategy. The adversarial task, i.e. \textit{center}, is marked by an overline.}
    \label{tab:panext}
    \begin{tabular}{c|c|c|c|c|c|c|c}
    \hline
Model &ID&main & area & count & contrast & $\overline{center}$ & {PanExt} \\\hline
\multirow{2}{*}{IV3}&1 & \checkmark & & & & & {$0.822 {\pm 0.01}$} \\
&8&\checkmark & \checkmark & \checkmark & \checkmark & \checkmark  & {$\textbf{0.847} {\pm \textbf{0,02}}$} \\\hline
\multirow{2}{*}{ResNet} & {R1} & \checkmark & & & & & {$0.800 {\pm 0.001}$} 
\\
& {R8}&\checkmark & \checkmark & \checkmark & \checkmark & \checkmark & {$\textbf{{0.895} }{\pm\textbf{{0.006}}}$} 
\end{tabular} 
\end{table}
\subsection{Sanity checks}
To confirm the benefit of the added related tasks, we compare these results with those obtained with random noise as additional targets.
This experiment is performed as a sanity check, where an auxiliary task is trained to predict random values. 
As expected, the overall, internal and external avg. AUCs are lower for this experiment and have larger standard deviations (overall avg. AUC $0.819 \pm 0.04$, int. test AUC $0.834 \pm 0.001$ and ext. avg. AUC $0.879 \pm 0.03$).
This shows that the selected tasks are more relevant to the main task than the regression of random values. 
\subsection{Interpretability and visualizations}
At this point, one may ask whether { it is possible to interpret the internal representation of the model to verify that the additional tasks were learned by the guided architectures. Table~\ref{tab:concept-performance} evaluates how well the representations of nuclei area, count and contrast are learned by the baseline and the proposed CNNs.} 
For model-ID3 (trained with the uncertainty-based weighting strategy), the prediction of the nuclei \textit{count} values has average determination coefficient $R^2 = 0.81 \pm 0.05$, showing that the concept was learned during training, passing from an initial Mean Squared Error (MSE) of the prediction of $0.46$ to $0.17$ at the end of training. Similar results apply to the other model-IDs 2 to 4 when only a single branch is added. Table~\ref{tab:concept-performance} compares the performance on the extra-tasks to learning the concepts directly on the baseline model activations, where the network parameters are not optimized to learn the extra tasks.
The classification of the \textit{center} in model-ID 5 reduces in accuracy as the gradient reversal is used during training.
The centers of the validation sets are predicted with accuracy $0.29 \pm 0.01$ at the end of the training (starting from  an initial accuracy of $0.53 \pm 0.01$). 
When more additional tasks are optimized together the performance on the side tasks is affected, with Model-IDs 6, 7 and 8 not reporting high $R^2$ values. The average $R^2$ of nuclei \textit{count} for model-ID 6, for example, decreases from $-2.25 \pm 0.05$ and plateaus at around $-0.63 \pm 0.05$. 
\begin{table}[!ht]
\fontsize{9}{13}\selectfont
    \centering
    \caption{Performance on the additional tasks for the baseline and guided models with the uncertainty-based strategy. The average and standard deviation of the determination coefficient are reported (the closer to $1$ the better).}
    \label{tab:concept-performance}
    \begin{tabular}{c|c|c|c}
    ID & area & count & contrast \\\hline
    baseline & $0.66 \pm 0.003$& $0.85 \pm 0.007$ & $0.56 \pm 0.01$ \\ \hline
    2 & $\textbf{0.70} \pm 0.005$ & - & -  \\\hline
    3 & - & $\textbf{0.88 }\pm 0.004$ & - \\\hline
    4 & - & - & $\textbf{0.64} \pm 0.003$ \\
    \end{tabular} 
\end{table}

\begin{figure}
    \centering
    \includegraphics[width=\linewidth]{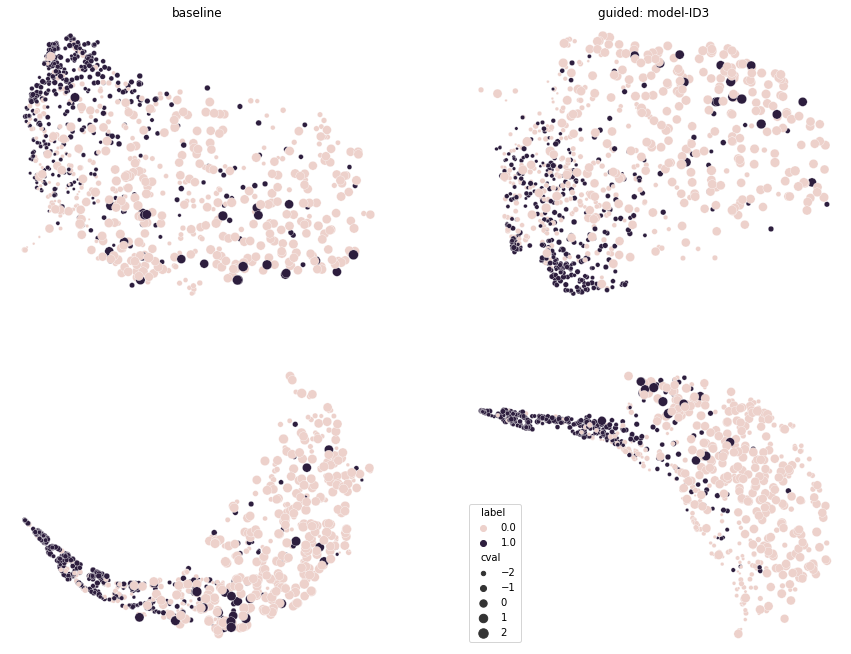}
    \caption{Uniform Manifold Approximation and Projection (UMAP) representation of the internal activations of the baseline and guided model-ID3 (obtained with the UMAP default hyperparameter set up). The top row shows the activations at the last convolutional layer of both models, known as mixed10 in the standard implementation of InceptionV3 (\cite{szegedy2016rethinking}). The bottom row shows the activations of the first fully connected layer after the GAP operation.}
    \label{fig:umap}
\end{figure}

Figure~\ref{fig:umap} shows the dimensionality reduction of the internal representations learned by the baseline and model-ID 3. The visualization is obtained by applying the Uniform Manifold Approximation and Projection (UMAP) method by \cite{mcinnes2018umap} (the hyper-parameters for the visualization were kept to the default values of 15 neighbors, 0.1 minimum distance and local connectivity of 1). The model-ID 3 selected for visualization was trained with the uncertainty-based weighting strategy. In the representation, the two classes are represented with different colors, whereas the size of the points in the plot is indicative of the values of nuclei counts in the images. 
The top row shows the projection of the internal representation of the last convolutional layer (known as mixed10 in the standard InceptionV3 implementation) of the two models. The bottom row shows the projection of the first fully connected layer after the GAP operation. 
Since the nuclei count values were normalized to zero mean and unit variance, these are represented in the plot as ranging between a minimum of -2 and a maximum of 2. For clarity of the representation, the image shows the UMAP of a random sampling of 4000 input images. 
\section{Discussion}
\label{sec:discussion}
The central question of this work is whether expert-knowledge can be used as a guidance to induce the learning of robust representations that generalize better to new data than the classic training of CNNs. The proposed experiments give multiple insights on this question that we discuss in this section.  

The clinical features used for diagnosis can be modeled as auxiliary and adversarial tasks. The extra tasks are modeled as regression tasks. This approach favors model transparency because it ensures that specific features of the data are learned in the compressed representation used for solving the main task. 
The features \textit{area} and \textit{contrast}, for example, were already modeled by \cite{graziani2018regression} as linear regression tasks that were used to probe the internal activations of InceptionV3 fine-tuned on the Camelyon data. These features emerged as relevant concepts learned by the network to drive the classification.
The architecture in this paper further guides the training towards learning a predictive relationship for these concepts. This is obtained by jointly optimizing the extra regression tasks together with the main task, encouraging the attention of the CNN on these aspects through multi-task learning even further \citep{caruana1997multitask}. 
{From the initial analysis of the baseline models in Table~\ref{tab:results}, the generalization performance of the Inception V3 baseline model (AUC $0.868$ on CamExt and $0.822$ on PanExt) is higher than the one obtained with ResNet (AUC $0.821$ on CamExt and $0.800$ on PanExt).}
All performances improve considerably even when a single extra task is added to the training, e.g. model-ID3. 
For this model, the representations of the positive class organize in a more compact cluster than in the baseline model, as shown by the UMAP visualization in Figure~\ref{fig:umap}. The representations on the right side of the figure (for model-ID3) also appear more structured than those on the left, being organized as following a direction for increasing values of the nuclei count.
With the feature values being extracted automatically, the modification of the Inception V3 model into a multi-task adversarial architecture does not require supplementary annotations, and only introduces a neglectable increase in complexity. One additional task, for instance, requires the training of only 2049 additional parameters, namely 0.008\% of Inception V3. 

The auxiliary and adversarial tasks are balanced in the same end-to-end training without additional tuning of the loss weight nor of a specific training schedule that would help the convergence of the adversarial task. This novel approach exploits the benefits of another paper in the machine learning research field that uses task-dependent uncertainty to structurally balance different losses such as MSE and BCE \citep{kendall2018multi}.
{  By learning the uncertainty as an additional parameter, the optimal loss weighting is directly found during training. The uncertainty estimated by our method is only the homoscedastic, task-dependent uncertainty. This is inherent to the task type and does not correspond to the observed, data-dependent main task uncertainty.}
{The multi-task framework is sensitive to the choice of the auxiliary tasks. Tasks to include during training should be selected based on prior knowledge and existing observations on whether they may cooperate or compete towards learning the main task. The result of concept-based interpretabily analyses in~\cite{graziani2020concept} was informative about for us to choose the tasks that were included in this study. 
The structured framework for identifying cooperating and competing tasks suggested in \cite{standley2020tasks} may be used when the prior knowledge about each tasks is not sufficient to establish whether a concept should be encouraged or discouraged. 
It is important to notice that appropriate weighting is also fundamental to avoid that the additional tasks overtake the main task.} The vanilla weighting of the losses shows instability on unseen domains and poor performance on the external test set.
The uncertainty-based approach, conversely, is robust to data variability and consistent over random seed initializations for all model-IDs. The stability to data variability is shown by the performance on the external test sets and by the testing with bootstrapping. The consistency over seed reinitializations is shown by the small standard deviation of the AUC on both test sets. 
This gives insight on how to handle the multiple loss types for the multi-task modeling on histopathology tasks. 
With the uncertainty-based weighting strategy the architecture did not require any specific tuning of the loss weights, whereas a fine-tuning of the weighting parameters appears highly necessary in the vanilla approach, particularly for the combinations with more than one extra task (model-IDs 6, 7, 8). 
The manual fine-tuning of the loss weights in the vanilla approach may lead to the over-specification of the model to the specific requirements of the test data considered in this study. 
{These observations can easily be extended to WSI classification. Patch-wise predictions, for instance, can be aggregated by the attention mechanism in~\cite{lu2021data} to obtain a slide-label. In this case, it may be interesting to further extend our work to study the impact of multi-task adversarial learning with slide-level concepts as additional tasks. 
}
These results not only extend the preliminary work by~\cite{gamper2020multi} to a different histology tissue and model architecture, but also give more insights on how to handle multiple auxiliary losses and adversarial losses without requiring tedious tuning of hyper-parameters. 

{It is important to notice, moreover, that high-order combinations of tasks do not automatically result in better generalization than low-order ones. The expressive capacity of the feature extractor is, in fact, kept the same for all the model combinations. The optimal grouping of tasks is an active area of research itself, and the estimation methods in~\cite{standley2020tasks} may be used to predict how a specific task grouping will perform.}

{Finally, the multi-task learning architecture provides improved robustness to label noise as opposed to multi-modal training, where the extra-task labels are passed as input signals. 
In multi-task learning, the labels are only used to determine the target output values. 
This means that any noise in the labels only affects the computation of the loss as an additional regularization term, further reducing the risks of overfitting. 
This is particularly convenient to model the additional targets. Accurate labels may not always be available and they can replaced by weak labels. In multi-modal training, on the other hand, the noisy signals are passed as input and suffer from propagation and amplification during the training process, making the latter less robust to label noise~\cite{caruana1997multitask}. 
Moreover, the multi-task learning approach introduces robustness to missing inputs, since the additional task labels are not needed at inference time. Our model can still be used in case the additional task labels are not available, differently from multi-modal settings. }
\section{Conclusion}
We show how expert-knowledge can be used pro-actively during the training of CNNs to drive the representation learning process. 
Clinically relevant and easy-to-interpret features describing the visual inputs are introduced as additional tasks for the learning objective, significantly improving the robustness and generalization performance of the model.
From a design  perspective, our framework aligns ethically with the intent of not replacing humans, but rather making them part of the development of deep learning algorithms. 
{ The flexibility of our method to include arbitrary additional features as desired or undesired learning targets is a key asset of our approach. New patterns may be identified by interacting with clinicians or by future studies and subsequently added to our models by retraining.}
{ Our experiments focus on tumor detection, but other tasks may benefit from a similar architecture. Deepfake detection, for example, may be improved by the auxiliary task of classifying checkerboard artifacts~\citep{wang2020cnn}}. Human-computer interfaces may be designed to directly collect user-specific feedback. 
The additional tasks may be used as a weak supervision to extend the training data with unlabeled datasets at a marginal cost of some additional automatic processing such as the extraction of nuclei contours or texture features. 
One may argue that additional annotations may be required for other clinical features. This represents, however, only a minor limitation of this method since a few annotated images may already suffice to train the additional tasks.

A few limitations of our method require further work and analyses. 
Our analysis is restricted to uncertainty-based weighting strategies, although several approaches were proposed in the literature \citep{leang2020dynamic}. The results on \textit{center} do not show a marked improvement by the adversarial branch. This could be due to the fact that the acquisition centers were not annotated for the PanNuke dataset. An unsupervised domain adaptation approach such as the domain alignment layers proposed by \cite{carlucci2017just} may be used to discover this latent information. 
Depending on the application, a different loss weighting approach may be used for the adversarial task and other undesired control targets can also be included, such as rotation, scale and image compression methods. 
In addition, our experiments show that the auxiliary tasks become harder to learn when they are scaled up in number, with model-ID 8 having a lower $R^2$ for the regression of the individual features than those reported for model-IDs 2 to 5 in Table~\ref{tab:concept-performance}. 
As explained also by~\cite{caruana1997multitask}, the poor performance on the additional tasks is not necessarily a problem as long as these help with improving the model performance and generalization on unseen data. 
Further research is necessary to verify how this architecture may be improved to ensure high performance on all the additional tasks, while maintaining its transparency and complexity at similar levels.
In future work we will also focus on extracting additional features exclusively from unlabeled data and on introducing them during training as weak supervision. 
\section*{Data Availability}
The Camelyon data that support the findings of this study are available at \url{https://camelyon17.grand-challenge.org/Data/} as accessed in June 2023, with the DOI identifier of the paper \url{https://doi.org/10.1109/TMI.2018.2867350}. 
The PanNuke data are available at \url{https://warwick.ac.uk/fac/cross\_fac/tia/data/pannuke} (accessed in June 2023), paper DOI \url{https://doi.org/10.1007/978-3-030-23937-4_2}. 
\section*{Code Availability}
The code used for the experiments is available online for reproducibility on Github (\url{https://github.com/maragraziani/multitask_adversarial}) and Zenodo at \url{ https://doi.org/10.5281/zenodo.5243433} (accessed in June, 2023). 


\acks{This work is supported by the European Union in the Horizon 2020 program with the projects ExaMode (grant s825292) and AI4Media (grant 951911). Part of this work was also supported by the Sinergia grant CRSII5 193832. Finally, we acknowledge Niccolò Marini for his feedback and insights.}

%
\ethics{The work follows appropriate ethical standards in conducting research and writing the manuscript, following all applicable laws and regulations regarding treatment of animals or human subjects.}

\coi{We declare we do not have any conflicts of interest.}

\bibliography{main}





\end{document}